\documentclass[journal]{IEEEtran}  

\usepackage{graphics} 

\ifCLASSOPTIONcompsoc
 \usepackage[caption=false,font=normalsize,labelfont=sf,textfont=sf]{subfig}
\else
 \usepackage[caption=false,font=footnotesize]{subfig}
\fi

\usepackage{graphicx}
\usepackage{epsfig} 
\usepackage{amsmath} 
\usepackage{amssymb}  

\usepackage{cite}
\usepackage{booktabs}

\usepackage{xcolor,soul}
\usepackage{enumitem,kantlipsum}

\usepackage{multicol}
\usepackage{hhline}
\usepackage{float}

\begin{document}

\title{Real-Time Forecasting of Driver-Vehicle Dynamics on 3D Roads: a Deep-Learning Framework Leveraging Bayesian Optimisation}

\author{Luca~Paparusso$^*$, Stefano~Melzi, and Francesco~Braghin%
\thanks{This work has been submitted to Elsevier for possible publication. Copyright may be transferred without notice, after which this version may no longer be accessible.}
\thanks{The authors are with the Department of Mechanical Engineering, Politecnico di Milano, Milan, Italy}
\thanks{$^*$ Corresponding author. Via Privata Giuseppe La Masa, 1, 20156 Milano MI, Italy. \tt\footnotesize luca.paparusso@polimi.it}%
}

\maketitle

\begin{abstract}
Most state-of-the-art works in trajectory forecasting for automotive target predicting the pose and orientation of the agents in the scene. This represents a particularly useful problem, for instance in autonomous driving, but it does not cover a spectrum of applications in control and simulation that require information on vehicle dynamics features other than pose and orientation. Also, multi-step dynamic simulation of complex multibody models does not seem to be a viable solution for real-time long-term prediction, due to the high computational time required.
To bridge this gap, we present a deep-learning framework to model and predict the evolution of the coupled driver-vehicle system dynamics jointly on a complex road geometry. It consists of two components. The first, a neural network predictor, is based on Long Short-Term Memory autoencoders and fuses the information on the road geometry and the past driver-vehicle system dynamics to produce context-aware predictions. The second, a Bayesian optimiser, is proposed to tune some significant hyperparameters of the network. These govern the network complexity, as well as the features importance.  The result is a self-tunable framework with real-time applicability, which allows the user to specify the features of interest.
The approach has been validated with a case study centered on motion cueing algorithms, using a dataset collected during test sessions of a \mbox{non-professional} driver on a dynamic driving simulator. 
A 3D track with complex geometry has been employed as driving environment to render the prediction task challenging. 
Finally, the robustness of the neural network to changes in the driver and track was investigated to set guidelines for future works.
\end{abstract}


\section{Introduction}

\begin{figure}[t]
\centering
\includegraphics[width=0.99\columnwidth]{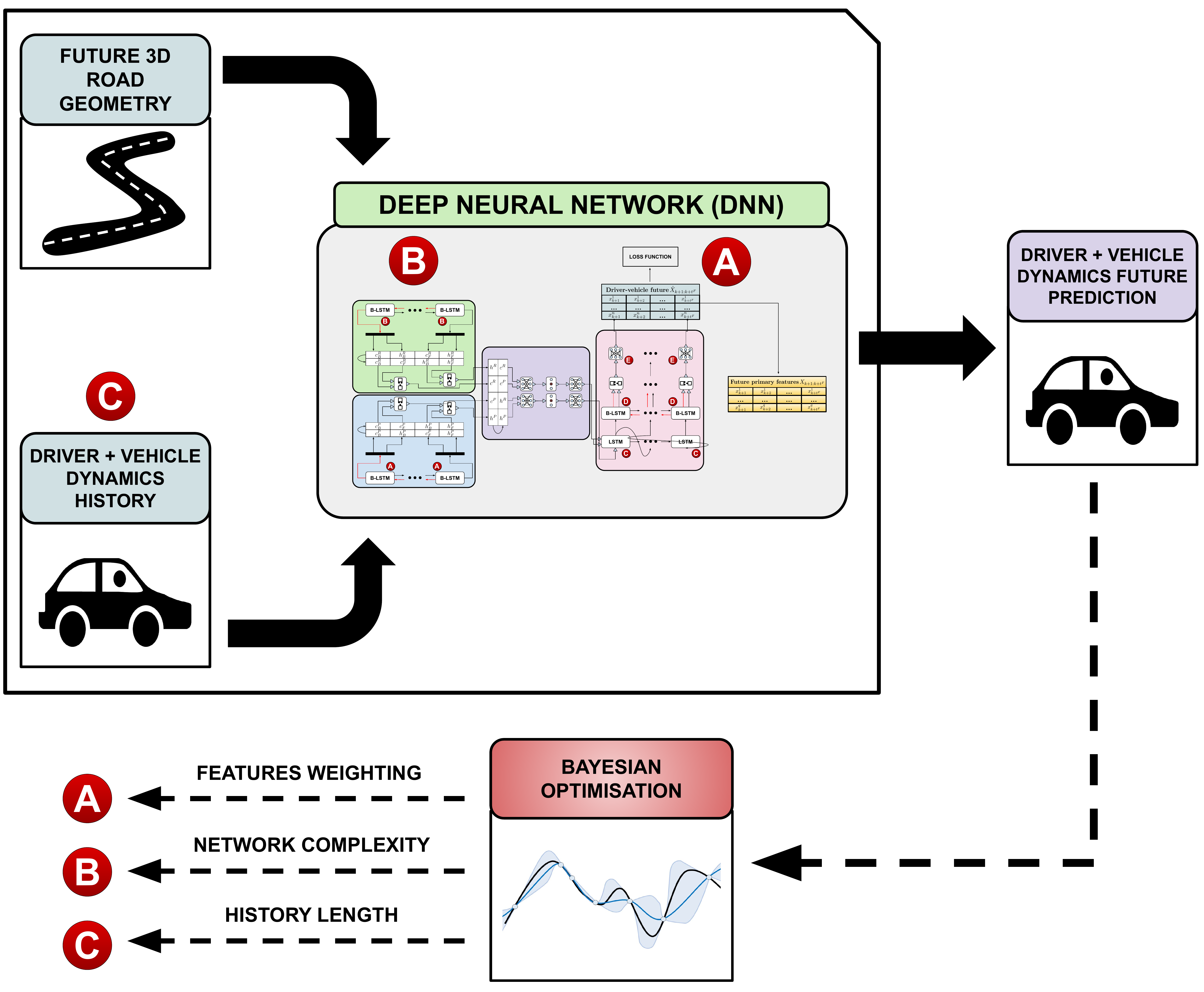}
\caption{Schematic representation of the proposed method. A deep neural network is used to forecast the future dynamics of the driver-vehicle system according to user-specified preferences. A 3D road geometry is considered to render the framework context-aware, guaranteeing generalisation on new roads and drivers with few training data. Finally, a Bayesian optimiser monitors the neural network outputs and tunes the prediction framework to guarantee good prediction performance as well as feasible computational times for online applicability.}
\label{fig: Proposed method}
\end{figure}

\IEEEPARstart{T}{he} field of control systems has considerably benefited from the advances in deep learning obtained in the last decade. 
The reason is to be found in the excellent results achieved by deep-learning algorithms in modelling complex dynamical systems \cite{kutz_deep_2017, yeung_learning_2019}. 
Particularly, modern control applications as autonomous driving and robot navigation, which deal with a dynamic and varied environment, need to manage more information to properly model the whole dynamics of the scene, including the behaviour of other autonomous agents or the interactions between humans and machines. 
As a consequence, the resulting dynamics becomes rather articulated, and in many cases it is even arduous to interpret the underlying physics that governs the system evolution. 
Under these conditions, data-driven models, especially deep-learning frameworks, have been demonstrated to embody a viable and efficient alternative to classical first-principle models.

Trajectory forecasting algorithms constitute one of the most important applications of deep learning in the area of control systems.
They generate predictions on the future motion of the agents in the control scenario, which can be used to enhance the promptness and performance of the controlled system. 
However, the majority of the state-of-the-art works in trajectory forecasting target predicting the future pose of the agents in the scene, that is their position and orientation in space. 
This is particularly useful when the core task of the downstream planning and control stacks is pose dependent only, for instance collision avoidance tasks.

On the other hand, there exist many automotive applications for which pose prediction of the vehicle-driver system is not sufficient to design proper simulation and control frameworks. 
As two motivating examples, predicting a more varied set of vehicle signals, as accelerations and rotational velocities, may significantly impact on motion cueing algorithms for dynamic driving simulators \cite{dagdelen_model-based_2009} and emergency forecasting for shared autonomy \cite{schwarting_safe_2018}. Also, even if the driver commands are known, simulating the vehicle dynamics with accurate multibody models is computationally expensive, so that long-term forecasts are not real-time obtainable.  
The extension of state-of-the-art trajectory forecasting algorithms to cope with a more varied set of signals is not trivial. 
In fact, it involves modelling and forecasting the dynamics of the coupled system constituted by a human driver and the vehicle that he/she drives at once, which has been a core problem in automotive research for half a century \cite{cacciabue_modelling_2007, plochl_driver_2007}.  
Supported by the previous motivating examples, in this work we aim to extend trajectory forecasting methods to design a simulation/prediction framework that is \vspace*{1mm}
\begin{itemize}
    \item capable to model and predict the main dynamics of the driver-vehicle system on a complex 3D track at once, that is forecasting the driver behaviour and modelling the vehicle response; \vspace*{1mm}
    \item \textit{data-driven}, to be tailored on a specific human driver;  \vspace*{1mm}
    \item \textit{road-geometry dependent}, to require few road geometry examples during training for good generalisation on new roads; \vspace*{1mm}
    \item \textit{computationally efficient}, to be employed in simulation and for online control tasks; \vspace*{1mm}
    \item \textit{self-tunable} during training, to adapt to specific applications and user-demanded output signals.
\end{itemize}

\noindent \textbf{Statement of contributions.} To bridge this gap, we propose a deep-learning framework based on LSTM autoencoders and leveraging Bayesian Optimisation, to predict a set of signals describing the driver-vehicle system dynamics. We call our framework \textit{DriVe-forecast}. The LSTM encoder-decoder structure is justified by many state-of-the-art works in trajectory forecasting. The following contributions are then added to cope with the above-stated requirements.
\begin{enumerate}
    \item The framework is designed to be general purpose, so that the user can specify the dynamics of interest. 
    Without losing generality, we propose a basic set of features that can be measured/estimated and used for common control applications, and that well represent the main driver-vehicle dynamics.
    \item Instead of splitting the problem into two parts, namely driver's intent prediction and vehicle dynamics simulation, the driver-vehicle dynamics is predicted at once. This allows to consider a coarser prediction horizon discretisation, which makes the computational time suited for real-time application. By splitting the problem instead, a reliable vehicle dynamics simulation would require a finer prediction horizon for the driver's intent, as well as a slower dynamics propagation by means of the multibody model.   
    \item The network fuses information on the realised past with the road geometry that the driver is about to travel, to generate context-aware predictions of the desired features for a predefined number of future time steps. 
    Differently from state-of-the-art methods, we consider a complex 3D road instead of a urban structured road (e.g. intersection, crossing) in the formulation. 
    This is done to model race tracks and suburban roads, in view of the applications mentioned in the previous section. However, we remark that the problem formulation can be integrated to support urban structured roads. 
    \item We introduce a parametric loss function to train the neural network. It allows to control the relative importance between forecasting errors on primary features, i.e. the ones of interest, and on secondary features, i.e. the auxiliary ones employed to help the network learning the dynamics of the system. Resorting to Bayesian optimisation, the loss function hyperparameters and the number of trainable parameters of the network are tuned to achieve a suitable trade-off between forecasting performance and computational time at the test phase. The latter is in fact fundamental in online implementations for control purposes. 
    \item The proposed approach is validated through a case study on motion cueing. An experimental campaign is conducted on a dynamic driving simulator to collect the dataset, train and validate the neural network. The dataset has been composed as follows.
    \begin{itemize}
        \item It refers to laps of a specific non-professional driver on a specific track. The driver is non-professional, so that each lap is different from previous ones. This makes the prediction problem more challenging, even considering testing on the same driver and track;
        \item we also include in the test portion of the dataset some laps performed by a different driver or on a different track to show the generalisation properties of the network, despite the constant driver and track kept during training.
    \end{itemize}
     The scope of this work is not to capture the behaviour of different drivers, but to link the predicted driver/vehicle dynamics (not only trajectories) to the road information, with limited computational burden. 
\end{enumerate}
A schematic representation of the proposed prediction framework is depicted in Figure \ref{fig: Proposed method}.

\noindent \textbf{Sections organisation.} The remainder of this paper is organised as follows. Related works on trajectory forecasting, vehicle dynamics modelling and motion cueing are presented in Sec. \ref{sec: related works}. The prediction problem is formulated in Sec.\,\ref{section: PROBLEM FORMULATION}. In Sec.\,\ref{sec: PROPOSED PREDICTION STRATEGY}, the proposed deep-learning architecture to solve the prediction problem is presented. The application of our method to the case study, centered on motion cueing algorithms, and the corresponding results are reported in Sec.\,\ref{section: APPLICATION TO A CASE STUDY}; in the same section, the dataset preparation and the hyperparameters tuning procedure are extensively described. In Sec.\,\ref{section: GENERALISATION ON NEW DRIVERS AND TRACKS}, we further investigate the generalisation properties of the proposed method, by testing it on data obtained with a different driver and a different track. Finally, the conclusions of this work and potential future studies are presented in Sec.\,\ref{section: CONCLUSIONS}.

\section{Related work}  
\label{sec: related works}
\subsection{Trajectory forecasting methods}
Predicting the future spatial location of agents in urban or structured road with deep learning has been an active research field in the recent years \cite{mozaffari_deep_2020, rudenko_human_2020}. In \cite{morton_analysis_2017}, Long Short-Term Memory (LSTM) neural networks are used to predict the acceleration distributions of cars on highway, modelling in-lane motion only. A probabilistic deep neural network is designed in \cite{ zyner_naturalistic_2020} to predict distributions of vehicles trajectories in a roundabout, i.e. in-plane positions at any time instant, assuming that each vehicle can be modelled through its in-plane position, heading and longitudinal velocity. In \cite{wang_deep_2018, zhang_recurrent_2020}, motion recognition and trajectory forecasting algorithms are applied to human-robot collaboration tasks, in which collision avoidance and safety guarantees constitute a strict requirement. Convolutional Neural Networks (CNN) are employed in \cite{dominguez_pedestrian_2017} to predict the motion of pedestrians for autonomous driving applications. 

Recent trends in multi-agent trajectory forecasting are geared towards conditioning the prediction of the surrounding agents pose on the ego-vehicle controls, to produce interaction-aware forecasts \cite{vedaldi_trajectron_2020}. A different approach to trajectory forecasting for control tasks, presented in \cite{ivanovic_mats_2020}, consists in producing interaction-aware state-space equation models instead of tracklets as outputs of the network, which can be directly employed for planning and control tasks.    

\subsection{Vehicle dynamics modelling}
The dynamics of the vehicle, in terms of response of its constitutive elements to known driving inputs, e.g. throttle, was widely investigated in the past \cite{gillespie_fundamentals_1992, milliken_race_1994, blundell_multibody_2004}. However, modelling the decision-making process of a human driver, namely the control actions that he/she executes considering environment, still represents an open research field. Unlike autonomous navigation systems, a human driver shows a varied, stochastic and, most of the times, non-optimal behaviour. Moreover, each driver has a personal driving style, which is strictly related to his/her own experience.

Many works have tried to emulate human drivers using control systems, e.g. \cite{braghin_race_2008}, with simplified models for the vehicle dynamics. These architectures allow to accurately describe the behaviour of professional drivers on track, whose aim is to minimise the lap time, but it could be difficult or even impossible to tune them so as to replicate a \mbox{non-professional} driver. To take into account the complex coupling effects between the driver and vehicle systems, and to capture the human nature of the driver, data-driven approaches may be employed to model and forecast the evolution of the whole system. Also, the high modelling performance of these methods allows to investigate how the geometry of the road influences the future driver/vehicle dynamics, which is considered in our work. 

\subsection{Motion cueing}
We also provide references to motion cueing, as it embodies one of the motivating examples of this work, and core of the case study for the evaluation of the performance of our method. 

Motion cueing is a planning algorithm that provides motion references to the dynamic platform of driving simulators \cite{asadi_robust_2017, mohammadi_multiobjective_2019}. 
The objective of the planning problem is to reproduce the translational accelerations and rotational velocities at the driver's head that would be generated on a real vehicle. 
In fact, it is through these quantities that humans perceive motion, thanks to their vestibular system. Modern implementations of motion cueing algorithms are based on constrained optimal control, to account for workspace limitations. 
The most used technique is MPC \cite{beghi_real_2012, beghi_real-time_2013, bruschetta_real-time_2017}, whose formulation includes the system and reference evolution in the future (predictive horizon) to generate an optimised and prompt motion, anticipating future situations. 
However, the state-of-the-art solutions developed so far do not fully exploit the potential benefits introduced by MPC. 
To avoid predicting the future behaviour of the driver, they employ a constant reference throughout the predictive horizon, equal to the one that is measured in simulation at the time of execution. 
As shown in \cite{grottoli_objective_2018}, knowing the reference future evolution in advance and extending the predictive horizon would allow to outperform the state-of-the-art results. This represents a motivating example for our work.

\section{Problem formulation}
\label{section: PROBLEM FORMULATION}
The formulation of the prediction problem shall be based on the emulation of the decision-making process of a human driver. 
At any time instant, a driver is aware of the vehicle state and his/her previous driving actions, which carry information about the recent past context that he/she has operated in. 
However, the past only does not provide a complete picture to characterise the resulting motion. 
In fact, drivers are highly conditioned by the road geometry that they see ahead, and act accordingly.
Therefore, the proposed formulation includes information on the road geometry into the prediction model, so as to highlight the dependency between driving actions and environment properties.

The previous statements are now formalised in mathematical notation. 
Let us consider the current time step $k \in \mathbb{N}$. Given the 3D geometry of the track, that is the spatial location of the centerline and the road margins;
given matrix $X_{(k-t^P:k)}$ describing the values assumed by $n \in \mathbb{N}$ vehicle features monitored in the last $t^P+1$ time steps, where $t^P \in \mathbb{N}$; 
the objective is to predict matrix $X_{(k+1:k+t^F)}$, describing the values assumed by $q \in \mathbb{N}$ vehicle features in the future $t^F$ time steps, where $t^F \in \mathbb{N},\,t^F\geq1$. 
The $q$ features are a subset of the $n$ features monitored in the past, thus $q \leq n$. 
The matrices $X_{(k-t^P:k)}$ and $X_{(k+1:k+t^F)}$ are defined as
\begin{equation}
\begin{aligned}
X_{(k-t^P:k)} := &
\left[
\begin{matrix}
x_{k-t^P}^1 & x_{k-t^P}^2 & ... & x_{k-t^P}^n\\[1ex]
x_{k-t^P+1}^1 & x_{k-t^P+1}^2 & ... & x_{k-t^P+1}^n\\[1ex]
... & ...& ...\\[1ex]
x_{k}^1 & x_{k}^2 & ... & x_{k}^n
\end{matrix}
\right],\\[1.5ex]
X_{(k+1:k+t^F)} := &
\left[
\begin{matrix}
x_{k+1}^1 & x_{k+1}^2 & ... & x_{k+1}^q\\[1ex]
x_{k+2}^1 & x_{k+2}^2 & ... & x_{k+2}^q\\[1ex]
... & ...& ...\\[1ex]
x_{k+t^F}^1 & x_{k+t^F}^2 & ... & x_{k+t^F}^q
\end{matrix}
\right],
\end{aligned}
\label{eq: matrices}
\end{equation}
where $x_i^j$ indicates the value of the $j$-th vehicle feature at the $i$-th time step. 
The 3D geometry of the track will be used to extract a set of $m \in \mathbb{N}$ road geometry features.

In the next sections, We detail the choice of the two main ingredients of the prediction framework, that is \vspace*{1mm}
\begin{itemize}
    \item a basic set of features describing the main driver-vehicle dynamics; \vspace*{1mm}
    \item a set of features encoding the future road geometry.
\end{itemize}

\begin{figure*}[t]
\centering
\includegraphics[width=0.99\textwidth]{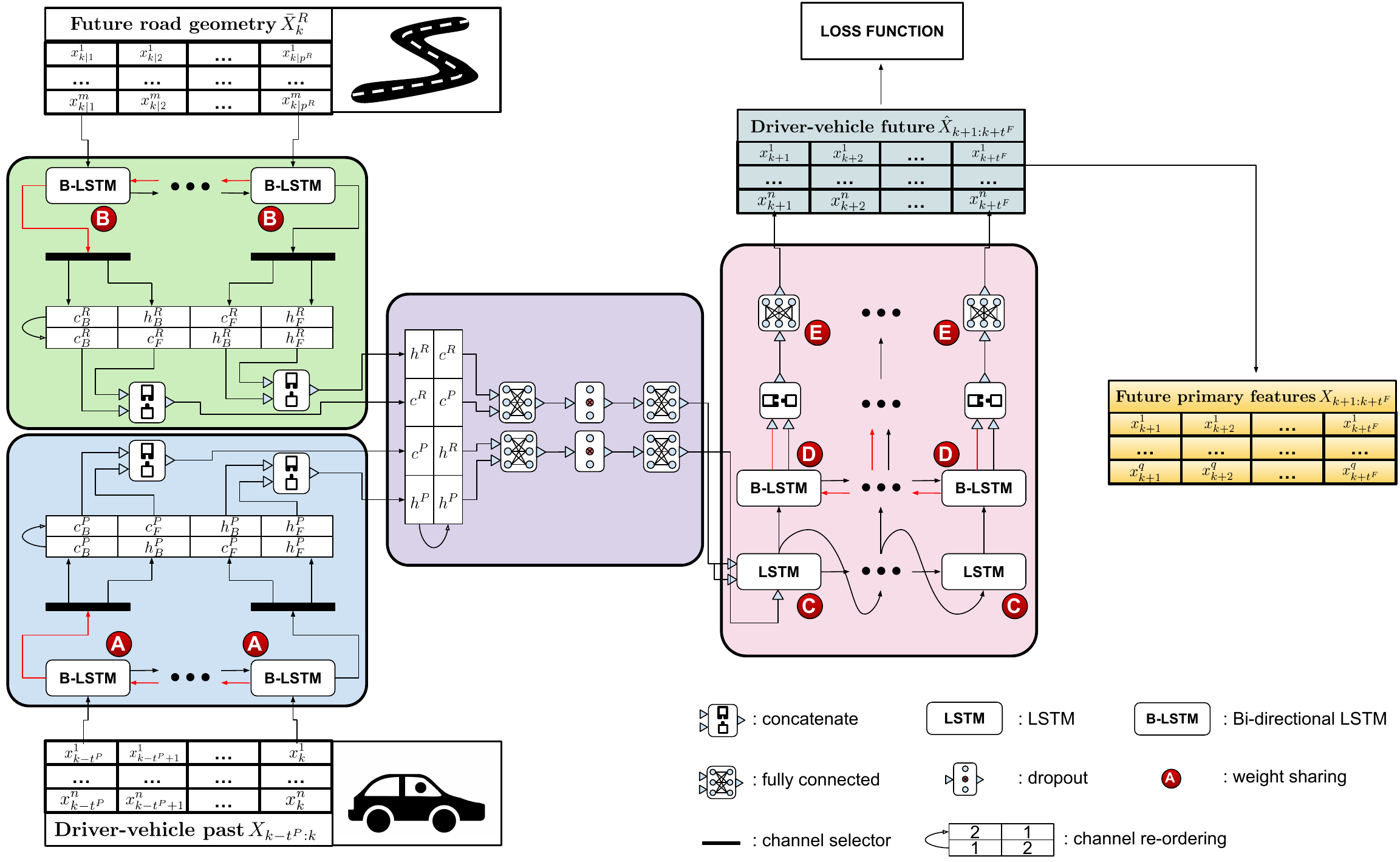}
\caption{Proposed deep neural network. The historical data (past) of the driver commands and vehicle states are encoded by a bi-directional LSTM to simultaneously represent the recent dynamics as well as the long-term context. In the same way, also the future road geometry is encoded by a bi-directional LSTM. The resulting cell states and hidden states of the two encoders are then passed through parallel fully connected layers, to fuse the past dynamics and future road information and generate the initial hidden and cell states for the decoder. The latter consists of an LSTM layer with recursion and a bi-directional LSTM. The output shape is finally obtained using fully connected layers.}
\label{fig: Network}
\end{figure*}

\subsection{Definition of the road geometry}
\label{subsection: Definition of the road geometry}
In prediction and control applications, it is fundamental to describe the geometrical properties of environment with features characterised by a sufficiently smooth evolution over time. 
To model the road and the vehicle position along the 3D road, we refer to the global coordinates of the left and right road margins, the road centerline and the vehicle centre of gravity (CG) with respect to a global orthonormal reference frame $X^G- Y^G-Z^G$.
The axis $Z^G$ points in the opposite direction with respect to the gravity vector.
We aim to derive parametric curves describing the margins and the centerline using $5^{th}$-order-polynomial splines, being the curvilinear abscissa the independent variable of each curve.
Starting from the sampled global coordinates of the road margins, the splines knots are placed using piecewise linear interpolation and resampling with a coarser spacing. 
Then, the splines parameters are computed by minimising the splines normed distance from the original sampled points.
The employment of $5^{th}$-order polynomials ensures the continuity of the splines up to the second derivative. 

Keeping the centerline curvilinear abscissa as the independent variable and using the splines, it is possible to derive the following road geometry features:
\begin{itemize}
    \item the road width;
    \item the first derivative of the coordinate in $Z_G$ with respect to the centerline curvilinear abscissa, i.e. the pitch slope;
    \item the road bank angle, i.e. the lateral slope;
    \item the centerline signed curvature in the $X^G-Y^G$ plane;
    \item the second derivative of the coordinate in $Z_G$ with respect to the centerline curvilinear abscissa.
\end{itemize}
The values of these features carry information on the context in which the vehicle operates.
The road width influences the lateral distance with respect to the centerline maintained by the driver along the track; 
the road pitch and lateral slopes play a key role in the dynamics of the vehicle, as they influence the orientation of the gravity vector with respect to the vehicle principal axes; 
finally, the centerline signed curvature in the $X^G-Y^G$ plane and the second derivative of $Z^G$ with respect to the centerline curvilinear abscissa provide information on the shape of the track, allowing to distinguish different types of turn and straightaway.

The global coordinates of the centerline are also employed as references to define the relative distance and relative yaw of the vehicle, which are included among the vehicle features, as described in the next section. 

\subsection{Definition of the vehicle features}
\label{subsection: Definition of the vehicle features}
As specified above, the $q$ vehicle features that we aim to forecast constitute a subset of the full set of $n$ vehicle features considered in the method.
The full set is fundamental to help the data-driven method learning the dynamics of the underlying physical process that we want to describe. 
Moreover, this becomes even more important considering that the driver could potentially lead the vehicle to very different situations on the 3D road, so that a data-driven algorithm failing to understand the driver-vehicle system dynamics would not be able to correctly generalise on new data. 
In view of these considerations, we will show in Sec.\,\ref{sec: PROPOSED PREDICTION STRATEGY} that the proposed data-driven scheme is designed to predict all of the $n$ features, while it adopts a weighting strategy to maximise the prediction performance on the $q$ desired features.

The selected $n$ vehicle features are:
\begin{itemize}
    \item the vehicle CG accelerations with respect to its principal axes;
    \item the vehicle chassis angular velocities with respect to its principal axes;
    \item the vehicle CG relative distance and vehicle chassis relative yaw with respect to the centerline;
    \item the vehicle CG velocities with respect to its principal axes;
    \item the throttle percentage, brake percentage, steering angle, steering angle rate and gear, i.e. driver commands.
\end{itemize}
In this case, the number of vehicle features is $n=16$. 
It is necessary to notice that the choice of the vehicle features depends on the level of detail in the description of the vehicle dynamics that one is interested to model. 
In this work, we decide to model the part of the driver-vehicle system dynamics constituted by those features that can be measured/estimated and used for common control applications, e.g. ADAS. 
However, the framework is designed to be general, so that the user can include other features of interest into the problem. The choice was also supported by a correlation analysis.

\section{Proposed prediction strategy}
\label{sec: PROPOSED PREDICTION STRATEGY}
The proposed prediction strategy is detailed in this section. The matrix $X_{(k-t^P:k)}$, introduced in Sec.\,\ref{section: PROBLEM FORMULATION}, will be referred to as matrix of the past from now on. It contains the information on the recent past that the system driver-vehicle has just experienced. As stated above, the past is not sufficient to fully characterise the behaviour of the system in the future, since it is highly correlated with the geometrical context that is encountered. To introduce the information on the road geometry that the driver sees ahead, we consider the curvilinear abscissa $s_k$ of point $P_k$, defined as the point of the centerline that is closest to the vehicle CG at the time step $k$. Considering that at the current time step the road geometry influencing the finite horizon prediction depends on a limited portion of the whole track, we define the maximum distance that the driver can see ahead $d^{R} \in \mathbb{R},\, d^R\geq 0$, and the number of spatially equidistant discretisation points $p^{R} \in \mathbb{N}$ in the continuous interval $(s_k,\, s_k+d^{R}]$. From the corresponding values of curvilinear abscissa, denoted as $s_{k|1},\,s_{k|2},\,...,\, s_{k|p^R}$, the road geometry features are evaluated and stored into matrix $\bar{X}^R_k$. Therefore, its structure is
\begin{equation}
\bar{X}^R_k :=
\left[
\begin{matrix}
x_{k|1}^1 & x_{k|1}^2 & ... & x_{k|1}^m\\[1ex]
x_{k|2}^1 & x_{k|2}^2 & ... & x_{k|2}^m\\[1ex]
... & ...& ...\\[1ex]
x_{k|p^R}^1 & x_{k|p^R}^2 & ... & x_{k|p^R}^m
\end{matrix}
\right],
\end{equation}
where $x_{k|i}^j$ indicates the value of the $j$-th road feature evaluated at the curvilinear abscissa $s_{k|i}$. 

Finally, following the considerations introduced in Sec.\,\ref{subsection: Definition of the vehicle features}, we introduce matrix $\bar{X}_{(k+1:k+t^F)}$ expressing the full set of $n$ vehicle features in the future $t^F$ time steps, as
\begin{equation}
\bar{X}_{(k+1:k+t^F)} :=
\left[
\begin{matrix}
x_{k+1}^1 & x_{k+1}^2 & ... & x_{k+1}^n\\[1ex]
x_{k+2}^1 & x_{k+2}^2 & ... & x_{k+2}^n\\[1ex]
... & ...& ...\\[1ex]
x_{k+t^F}^1 & x_{k+t^F}^2 & ... & x_{k+t^F}^n
\end{matrix}
\right].
\end{equation}
We remark that matrix $X_{(k+1:k+t^F)}$, i.e. the goal of our problem, can be obtained by extracting the subset of $q$ features of interest from matrix $\bar{X}_{(k+1:k+t^F)}$, being $q \leq n$.

\subsection{Deep neural network}
\label{sec: Proposed Neural Network structure}
To solve the prediction problem formulated above, we propose a deep-learning architecture. The structure of the artificial neural network, exploiting an encoder-decoder topology to synthesize the information of the past $X_{(k-t^P:k)}$ and the road seen ahead $\bar{X}^R_k$, is now presented.  

The scheme of the network is shown in Figure \ref{fig: Network}. 

The matrix of the past $X_{(k-t^P:k)}$ is encoded using a bidirectional Long-Short-Term-Memory (LSTM) layer with $u_{e}$ hidden units, for their high performance in sequence encoding \cite{britz_massive_2017}. The last hidden states $h_F^P, h_B^P$, and cell states $c_F^P, c_B^P$ of the LSTM layers in the forward and backward directions are merged together through concatenation, to form one hidden states vector $h^P$ and one cell states vector $c^P$. The road geometry matrix $\bar{X}^R_k$ is encoded by means of another bidirectional LSTM layer with $u_{e}$ hidden units, employing the same merge mechanism, to obtain the hidden states vector $h^R$ and cell states vector $c^R$. The hidden states and cell states of the two encoders are then concatenated, to create an extended hidden states vector $h$ and an extended cell states vector $c$. Two dense layers with $2u_{e}$ and $u_{d}$ neurons are applied to both $h$ and $c$ without weights sharing, to merge the information carried by the encodings of the past and the geometry of the road in the future, and to uniform them to the dimensions of the decoder. A dropout layer with rate $r$ is placed between each couple of dense layers to promote regularisation.  Consequently, the vectors $h^D$ and $c^D$ are generated. 

The decoder consists of two stages. The first is a unidirectional LSTM layer with $u_{d}$ hidden units, whose hidden and cell states are initialised using the vectors $h^D$ and $c^D$. The first cell is fed with $h^D$ as inputs. From the second cell on, the output of each LSTM cell is recursively used as input of the successive cell. The second stage of the decoder is a bidirectional LSTM layer of $u_{d}$ hidden units, whose output sequences are fed to a final time-distributed dense layer, that brings the matrices to assume the output shape. The output of the network $\hat{X}_{(k+1:k+t^F)}$ is the prediction of $\bar{X}_{(k+1:k+t^F)}$, which represents the ground truth. The entries of $\hat{X}_{(k+1:k+t^F)}$ are denoted with $\hat{x}_i^j$, being it the value of the $j$-th vehicle feature at the $i$-th time step.

Finally, we define the two hyperparameters $u_{e,d}$ and $\xi$ as
\begin{align}
    u_{e,d} =& u_{e} + u_{d},\\
    \xi =& \dfrac{u_{e}}{u_{e,d}}.
\end{align}
These two hyperparameters synthetically express the total network complexity for the encoding and decoding parts. Since the total network complexity influences the computational time needed to generate a prediction, it is important to properly limit the values of $u_{e,d}$ and $\xi$ to render the proposed architecture feasible for online control applications. This consideration will be reminded in Sec.\,\ref{sec: Training and validation of the model}, where hyperparameters tuning is described.

\subsection{Loss function}
\label{sec: Loss function}
Taking advantage of how the problem is formulated in Sec.\,\ref{section: PROBLEM FORMULATION}, the proposed prediction framework is potentially able to forecast the values of any vehicle feature that one could be interested in. Independently from the specific subset of $q$ features of interest for a determined application, it is important to design a loss function that penalises prediction errors on the full set of $n$ vehicle features. This choice forces the network to learn the driver-vehicle system dynamics, and infuses a natural generalisation into the data-driven approach. The $q$ features of interest and the remaining $n-q$ ones are denominated as primary and secondary features, respectively.

To comply with the requirements of specific applications, in which one could be interested in maximising the prediction performance on the restricted set of $q$ vehicle features, we employ Weighted Mean Square Error (WMSE) as loss function:
\begin{equation}
\mathcal{L} = \sum_{i=k+1}^{k+t^F} \dfrac{\sum_{j=1}^{n} w^j(\hat{x}_i^j - x_i^j)^2}{n},
\label{eq: loss function}
\end{equation}
where $w^j=1$ for $j=1,...,q$ are the weights on the primary features, and $w^j=w$ for $j=q+1,...,n$ are the weights on the secondary features. The hyperparameter $w$ is in the range $0\leq w \leq 1$, and is tuned depending on the application. The employment of the weighting term $w$ is strictly related to the requirements of online applicability of the prediction strategy. In case online applicability were not an issue, the network complexity could be arbitrarily increased to predict all of the $n$ vehicle features. Instead, in the case study of this work we aim to design a network with prediction times that are suitable for online control applications; thus, we use $w$ to help the network identify the relative priority between the primary and secondary features.  

\section{Experiments}
\label{section: APPLICATION TO A CASE STUDY}

\begin{figure}
\centering
       \includegraphics[width=0.90\columnwidth]{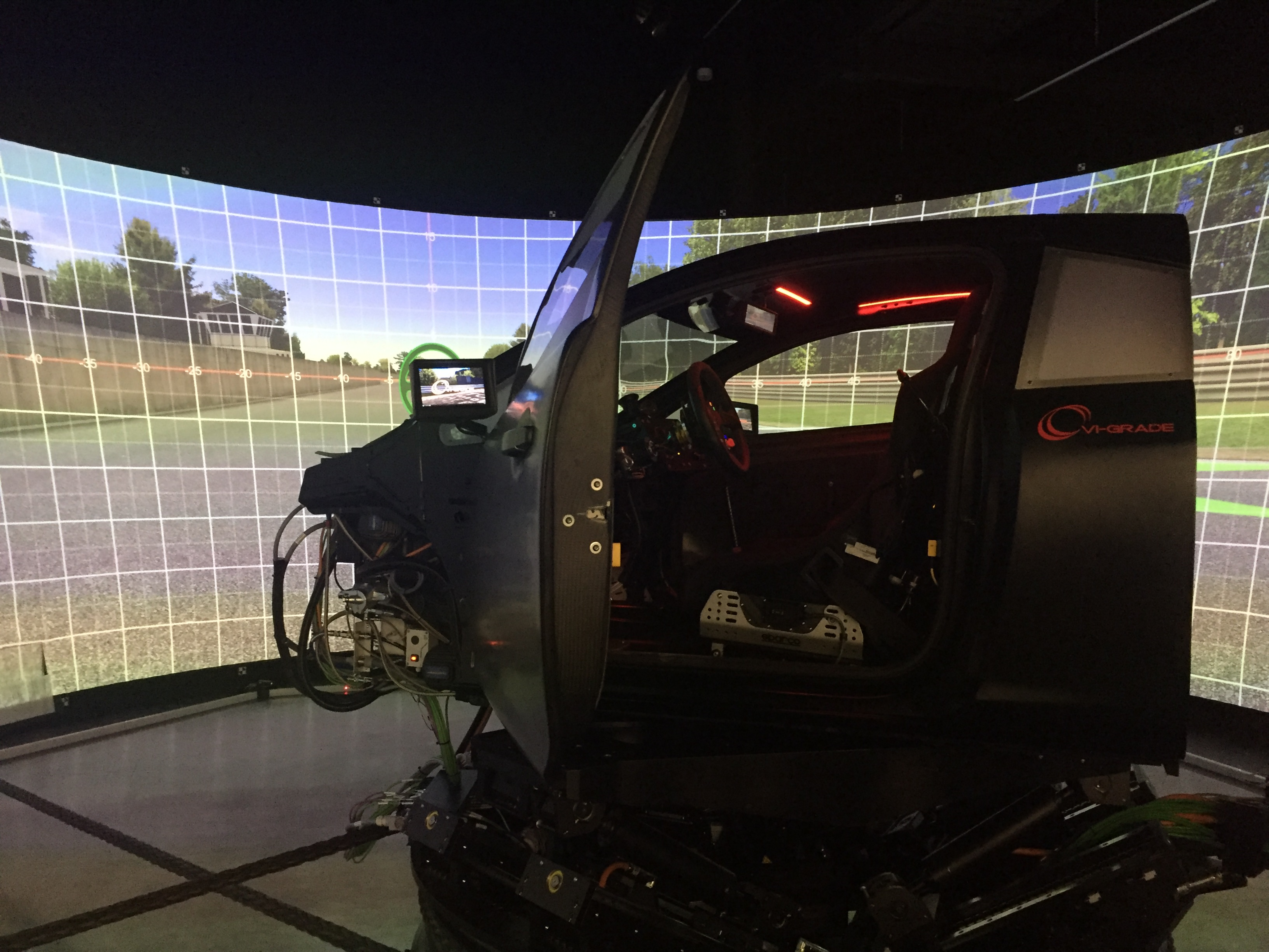}
\caption{Dynamic driving simulator used to collect the dataset.}
\label{fig: Driving simulator}
\end{figure}

\begin{figure}
\centering
       \includegraphics[width=0.90\columnwidth]{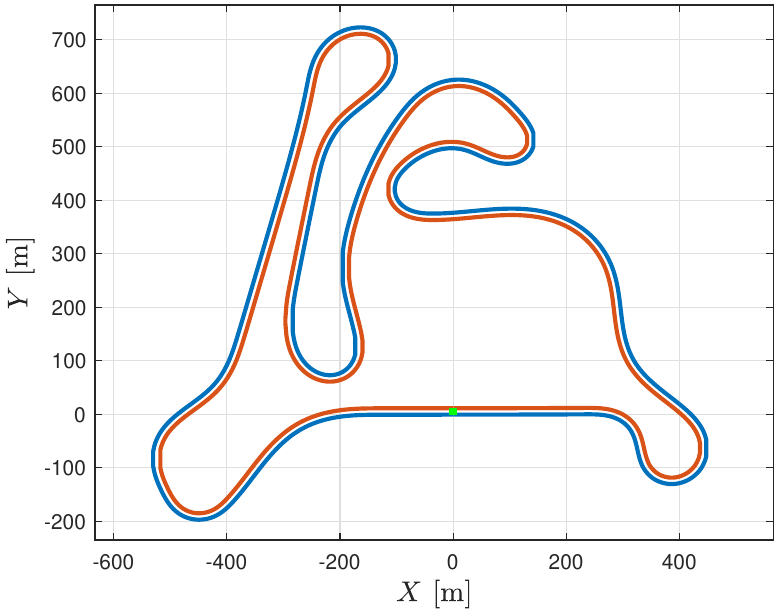}
\caption{Planimetry of the test track Calabogie used for the validation of the proposed approach. The right and left margins are shown in red and blue, respectively, i.e. the lap is clockwise. The start line is marked in green.}
\label{fig: Calabogie map}
\end{figure}

\begin{figure}
\centering
       \includegraphics[width=0.90\columnwidth]{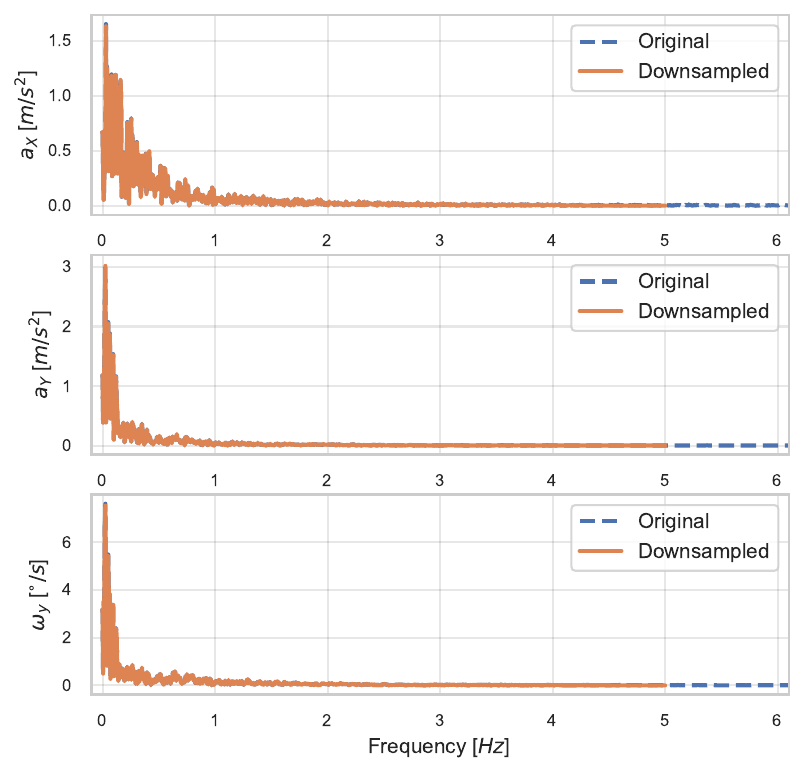}
\caption{Frequency content of the longitudinal acceleration $a_X$, lateral acceleration $a_Y$ and yaw rate $\omega_y$, for the original and downsampled signals. The downsampled signals contain approximately $98\%$ of the power of the original ones.}
\label{fig: Downsampled}
\end{figure}

The proposed framework is validated by applying it to a challenging case study, namely the prediction of the features that are commonly fed as references to any modern motion cueing algorithm for dynamic driving simulators based on Model Predictive Control (MPC). Our code and the dataset are online available\footnote{https://github.com/lpaparusso/DriVe-forecast}. We remark that the proposed methodology can also be employed to predict the evolution of any other feature that could be of interest for other automotive applications; we leave the study of further applications to future works.

As we remarked in Sec. \ref{sec: related works}, forecasting the vehicle accelerations and rotational velocities in driving simulators may significantly improve the simulation experience. For this reason, we will focus the attention of our network to prioritise the reconstruction of the vehicle CG longitudinal and lateral accelerations ($a_X$ and $a_Y$, respectively), and vehicle chassis yaw rate ($\omega_y$), being the in-plane motion the most crucial to enhance the simulator workspace utilisation. This means that the number of primary features is $q=3$. 
We consider a prediction horizon length $t^F=30$ (i.e. $3$\,s). According to the maximum vehicle speed, we estimate that a suitable choice for the hyperparameters that model the road geometry in the future is $d^R=150$\,m and $p^R=50$.  

\subsection{Dataset}
\label{sec: Dataset structure}
The dataset used to validate the prediction framework was generated on a dynamic driving simulator, shown in Figure\,\ref{fig: Driving simulator}. The simulation environment set up for our analysis is the Canadian test track Calabogie, whose planimetry is shown in Fig.\,\ref{fig: Calabogie map}. We collected a dataset of 36 laps performed by a \mbox{non-professional} driver, for a total of approximately $75$ minutes and $160$\,km travelled. A \mbox{non-professional} human driver was selected for the tests to introduce high variability into the resulting motion trajectories, which makes the prediction problem particularly challenging. The low level of repeatability of the driver at each lap is used to test the generalisation capability of the network. The latter is in fact expected to generate predictions that are close to the ground truth when the driving behaviour is more repeatable, and stabilise around an average behaviour when it is not.

The original data were acquired at a sampling frequency of $100$ Hz. To be able to consider long-enough prediction horizons while saving the neural network computational time, data are downsampled at a frequency of $10$ Hz, after using a zero-phase anti-aliasing filter with cut-off frequency at $4$ Hz. As an example, Figure\,\ref{fig: Downsampled} shows the original and downsampled frequency content of $a_X$, $a_Y$ and $\omega_y$. The downsampled signals contain approximately $98\%$ of the power of the original ones; thus, the downsampling architecture allows to obtain a good trade-off between the level of detail in the representation of the driver-vehicle system dynamics and the speed of the prediction algorithm, in terms of online computational time at the test phase. Lastly, the signals are normalised to facilitate the neural network training process.

\subsection{Training and validation of the model}
\label{sec: Training and validation of the model}
The matrices defined in \eqref{eq: matrices} are obtained by windowing each lap of the original dataset, using unitary shift and stride. A $31-4-1$ random split of the laps is applied to obtain the training, validation and test sets, respectively. 

The model is trained using the Adam optimiser \cite{kingma_adam_2014}, with $\beta_1=0.9$, $\beta_2=0.999$ and $\epsilon=10^{-7}$. After a first coarse tuning, it was observed that the hyperparameters affecting the network performance are the dropout rate $r$, the weight $w$ on the secondary features, the number of time steps of the past $t^P$, and the hyperparameters $u_{e,d}$ and $\xi$ defining the structure of the network. To fine tune the hyperparameters, we employ Bayesian optimisation \cite{pelikan_boa_1999} (using the toolbox \cite{bayesian_opt_toolbox}), which targets finding the best value of the metrics $\mathcal{M}$ on the validation set. The metrics $\mathcal{M}$ is the Mean Absolute Error (MAE) referred to the primary features only, i.e. $a_X$, $a_Y$ and $\omega_y$, 
\begin{equation}
\mathcal{M} = \sum_{i=k+1}^{k+t^F} \dfrac{\sum_{j=1}^{q} |\hat{x}_i^j - x_i^j|}{q},
\label{eq: metrics function}
\end{equation}
being $|\cdot|$ the absolute value operator. Bayesian optimisation implements $2$ steps of random exploration and $10$ iterations.

The training process of each model generated by Bayesian optimisation is performed in two phases. In the first phase, the Adam optimiser starts from a random initialisation of the weights of the neural network; to re-initialise the optimizer with weights that are closer to the optimal ones, which favours a more precise gradient descent, the second phase starts from the final configuration of the first phase. We now detail the characteristics of the two training phases. The first phase implements an exponentially decaying learning rate, going from $0.001$ to $0.0005$ in $34$ epochs; after epoch $34$, the learning rate is kept constant. We set the maximum number of epochs to $100$, while using an early stopping option that activates when the validation metrics does not decrease within $25$ consecutive epochs. The second phase employs a constant learning rate equal to $0.0001$, with the same early stopping option as the first phase and a maximum number of epochs equal to $500$. A batch size equal to $64$ showed the best performance.

Considering the high variability of motion of the \mbox{non-professional} driver and the relatively small size of the dataset, we perform cross-validation to obtain performance indices that do not depend on the specific data split. Therefore, fixing the optimal hyperparameters obtained through Bayesian optimisation, training is finally repeated $10$ times, changing the training, validation and test sets at each iteration. This allows to better analyse the framework performance and generate meaningful statistics. These values will be shown in the next sections. 

\subsection{Ablation experiment}
\label{sec: Analysis of the results}
An ablation experiment is carried out to certify the effectiveness of the neural network structure in generating context-aware predictions. In this regard, training is repeated removing the blocks of the network corresponding to the future road geometry. With reference to Figure \ref{fig: Network}, the green block is removed, and its outputs are treated as empty arrays.

The range of variation of each hyperparameter in Bayesian optimisation, and their optimised values for the original and ablated experiments, are reported in Table \ref{tab: Bayesian ranges}. We highlight that the ranges of the hyperparameters $u_{e,d}$ and $\xi$ are properly set to reduce the complexity of the neural network, thus allowing to use the prediction approach in online control applications. It is interesting to notice that there is correspondence among the optimised hyperparameters in the two cases. The optimal number of steps in the past is $t^P=37$ in both cases. This highlights that it is sufficient to observe $3.7$\,s of the past to characterise the current driving condition/situation, which is likely for the dynamics of the driver-vehicle system. The only hyperparameter that differs in the original and ablated experiments is the secondary features weight $w$. It is reasonable that the ablated experiment employs higher weights for the secondary features, as it cannot contextualise the dynamics according to the road. Thus, the ablated neural network has to focus more intensively on the relationship between all of the variables, i.e. the vehicle dynamics, to support future predictions.

\begin{table}
\caption{Range of the hyperparameters in Bayesian optimisation, and their optimised values in the original and ablated experiments.}
\centering
\begin{tabular}{|c||c|c|c|c|}
\hline
\, & \, & \, & \, & \,\\
\textbf{Hyperparameter} & \textbf{min} & \textbf{max} & \textbf{Ours} & \textbf{Ablated}\\
\textbf{\,} & \, & \, & \, & \,\\
\hhline{|=||=|=|=|=|}
\, & \, & \, & \, & \,\\
Dropout rate $r$ & 0.25 & 0.5 & 0.25 & 0.27\\[0.5ex]
Secondary features weight $w$ & 0.0 & 1.0 & \textbf{0.23} & 0.40\\[0.5ex]
Past horizon $t^P$ & 15 & 40 & 37 & 37\\[0.5ex]
Network complexity $u_{e,d}$ & 70 & 110 & 97 & 94\\[0.5ex]
Encoder relative complexity $\xi$ & 0.3 & 0.5 & 0.31 & 0.39\\[1ex]
\hline
\end{tabular}
\label{tab: Bayesian ranges}
\end{table}

\begin{table*}
\caption{Sample average $\mu$ and standard deviation $\sigma$ of the loss and metrics using cross-validation. The values are referred to the normalised data. We remark that, since \eqref{eq: loss function} and \eqref{eq: metrics function} refer to a single window of prediction, the loss/metrics over the full sets is obtained by summing up each loss/metrics and dividing by the total number of windows.}
\centering
\begin{tabular}{|c||c|c|c|c|}
\hline
\, & \multicolumn{2}{c|}{\,} & \multicolumn{2}{c|}{\,}\\
\textbf{Sets} & \multicolumn{2}{c|}{$\mathbf{Ours}$} & \multicolumn{2}{c|}{$\mathbf{Ablated}$}\\
\, & \multicolumn{2}{c|}{\,} & \multicolumn{2}{c|}{\,}\\
\cline{2-3} \cline{4-5}
\, & Loss & Metrics & Loss & Metrics\\
\cline{2-3} \cline{4-5}
\hhline{|=||=|=|=|=|}
\, & \, & \, & \, & \,\\
Training & $0.0130 \pm 0.0013$ & $0.1158 \pm 0.0077$ & $0.0275 \pm 0.0019$ & $0.1201 \pm 0.0054$\\[0.5ex]
Validation & $0.0223 \pm 0.0015$ & $\mathbf{0.1572 \pm 0.0059}$ & $0.0461 \pm 0.0038$ & $0.1668 \pm 0.0079$\\[1ex]
\hline
\end{tabular}
\label{tab: Bayesian loss-metrics}
\end{table*}

\begin{figure*}[t]
\centering
       \includegraphics[width=0.90\textwidth]{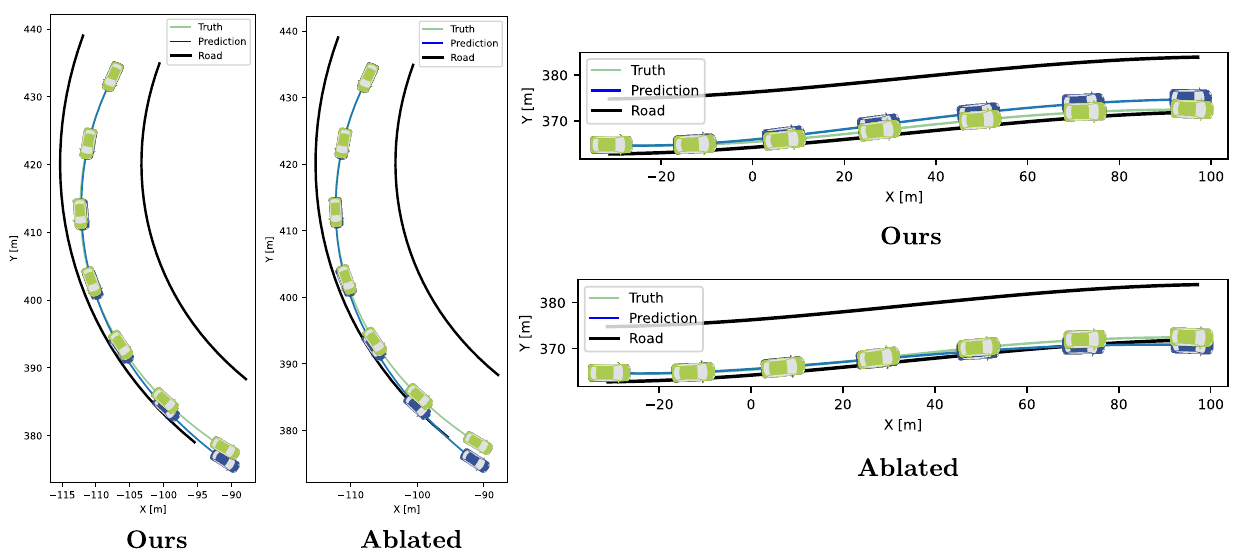}
\caption{Simulated vehicle trajectories obtained by integration of the predicted primary features. The ablated experiment (i.e. the one without the road geometry encoder), forecasts trajectories that overcome the road margins. On the contrary, the complete model helps maintaining the vehicle trajectories within the road margins.}
\label{fig: Ablation}
\end{figure*}

\begin{figure*}[t!]
\centering
 \includegraphics[width=0.90\textwidth]{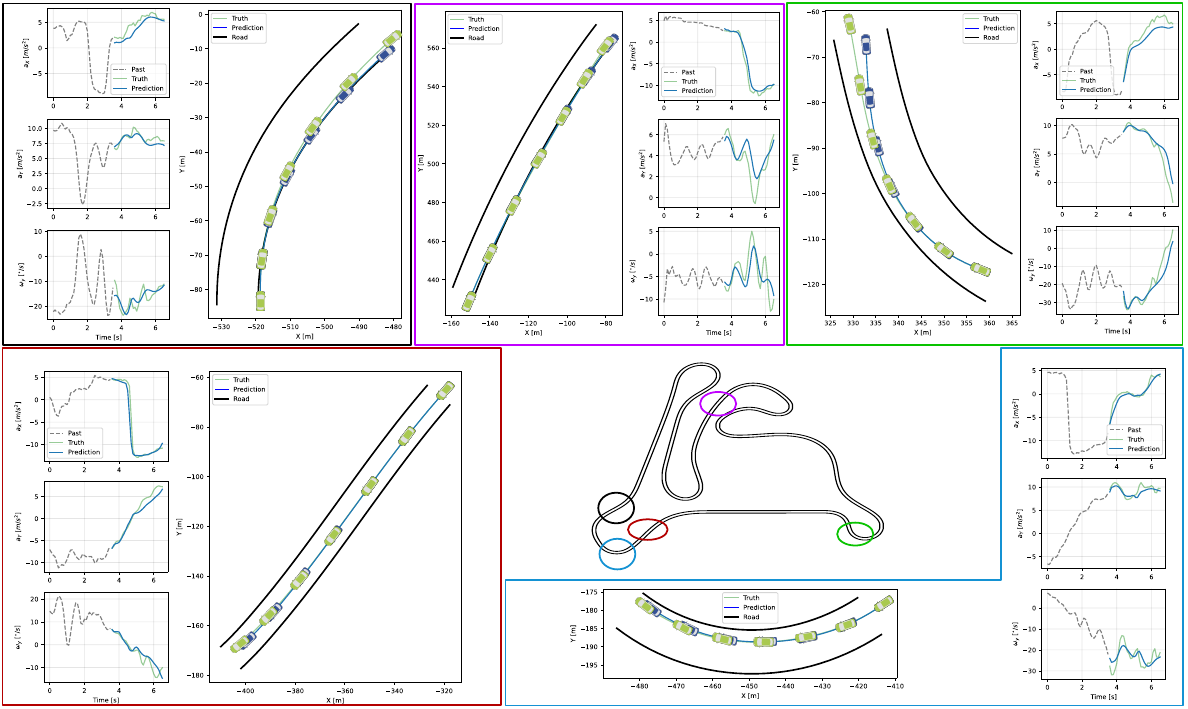}
\caption{Exemplary prediction horizons along the track extracted from the test set. The predictor shows good accuracy even at the latest time steps of the prediction horizons.}
\label{fig: test_ph}
\end{figure*}

The statistics of the loss and metrics for the $10$ cross-validation iterations are reported in Table \ref{tab: Bayesian loss-metrics}. As expected, the proposed model shows better performance on the validation set compared to the ablated model. 

Finally, to better show the efficacy of the full model against the ablated one, we compute the predicted vehicle pose and orientation by integration of the predicted primary features. The corresponding differential equations are obtained by means of simple kinematic considerations. As shown in Figure \ref{fig: Ablation}, the road geometry encoder helps keeping the vehicle trajectories within the road margins.

\subsection{Analysis of the model performance}
\begin{figure}
\centering
      \includegraphics[width=0.9\columnwidth]{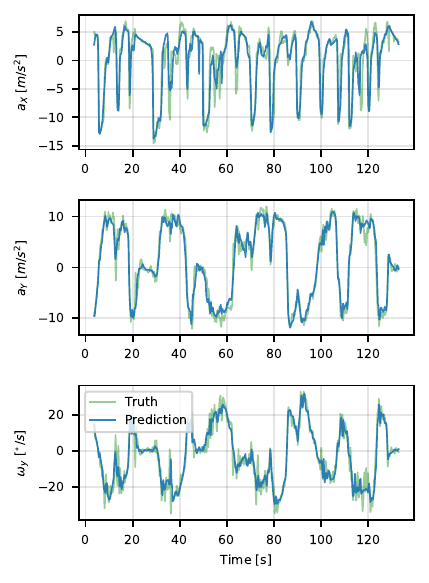}
\caption{Prediction of the primary features along the test lap. Each point of the orange curves represents the value predicted $30$ time steps ($3$\,s) in advance.}
\label{fig: testred_whole}
\end{figure}

The prediction of the variables of interest, i.e. $a_X$, $a_Y$ and $\omega_y$, along the test lap is reported in Figure \ref{fig: testred_whole} for the two prediction horizons. Despite the highly variable behaviour of the \mbox{non-professional} driver, the proposed prediction framework succeeds in distinguishing the different portions of the road and generates context-aware predictions. As expected, the forecasting performance is higher in the points of the track where the driver behaviour is more repeatable. Instead, the network filters out the fast oscillations that derive from sporadic actions, stabilising around an average value that still properly captures the dominant trend of the dynamics. 

To better visualise the forecasting performance in whole prediction horizons, we report some examples in Figure\,\ref{fig: test_ph}.

\begin{figure}[htbp]
\centering
     \includegraphics[width=0.92\columnwidth]{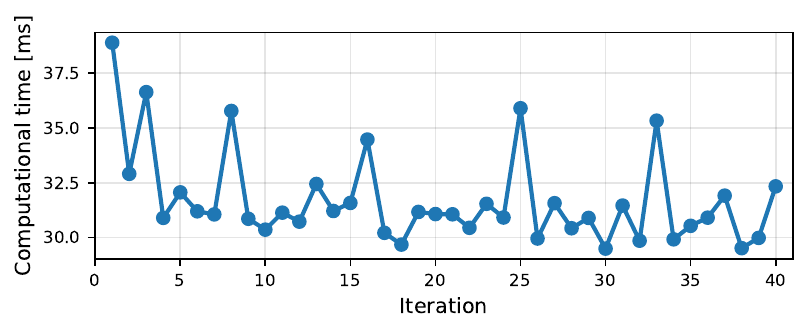}
\caption{Computational time of the neural network for $40$ consecutive prediction horizons at test time.}
\label{fig: ct}
\end{figure}

Finally, we evaluate the computational time of the neural network at the test phase. The results obtained for $40$ consecutive windows are shown in Fig. \ref{fig: ct}. Considering the neural network trained for $t^F=30$, the average computational time of a prediction window is $32$\,ms, with a sample standard deviation of $2$\,ms. 
The tests were run using interpreted Python code on a PC with an Intel Core i7-8565U CPU and 16 GB RAM. The measured computational times show that the proposed method is already suitable for simulation and online control applications. We remark that it may be possible to further increase performance by resorting to compiled code and executing the operations on dedicated GPUs.

We finally provide an example showing the reduction of computational time introduced by our method. The computational time required to run a multibody simulation, considering the same prediction horizon and known control inputs (straight acceleration), is around $500$\,ms on the same PC. Moreover, if the control inputs are unknown, as in the problem investigated in this work, the computational time would become even higher, due to serial computations. This example highlights the benefits introduced by our approach, which forecasts the driver-vehicle dynamics at once.

\section{Generalisation on new drivers and tracks}
\label{section: GENERALISATION ON NEW DRIVERS AND TRACKS}
Generalisation on the driver's behavioural variability, which is the core of this work, has already been investigated in the previous section. However, as an extra benefit of our approach, we show that the trained neural network can generalise with fair accuracy on new drivers and tracks, despite the limited amount of data seen at training time. In other words, the trained neural network with $t^F=30$ is used to predict state trajectories from a new dataset, generated under considerably different high-level conditions with respect to the ones employed for training and validation. In particular, we evaluate the prediction performance of the network by changing driver and track, respectively, in two separated tests. This check is meant to assess how far the current training/validation set is from an ideal one whose range of variability allows to achieve high prediction performance on a generic high-level scenario.

The new track is obtained by travelling the original test track in opposite direction. For completeness, it has to be remarked that a track travelled in opposite direction generates driver-vehicle dynamics that are totally different and independent from the ones experienced in the original direction.

\begin{figure}
\centering
\subfloat[][]{\includegraphics[width=0.90\columnwidth]{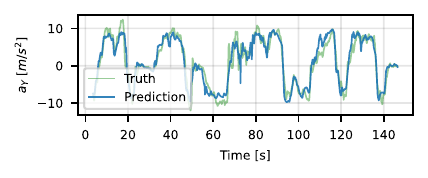}}\\
\subfloat[][]{\includegraphics[width=0.90\columnwidth]{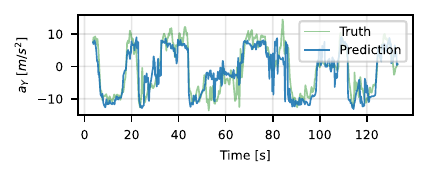}}
\caption{Prediction of the lateral acceleration, using a test lap performed by a different driver (a) and on a different track (b). Each point of the orange curves represents the value predicted $30$ time steps earlier.}
\label{fig: test_other}
\end{figure}

The prediction performance for the lateral acceleration $a_Y$ is shown in Figure \ref{fig: test_other} for the two cases. The network shows good generalisation on the two high-level variations. Particularly, it is interesting how our framework manages to reconstruct the main trend of the signal on the new track, despite the training set contained only data from a single track, which represent few of the possible combinations of road geometry. This confirms that the proposed architecture succeeds in linking the decisions of the driver with the geometry of the road. Thus, by selecting a suitable set of tracks to populate the dataset, it may be possible to optimise the generalisation capabilities of the network, making it robust to new tracks. This analysis is left to future works.

As an interesting future work, transfer learning \cite{tan_survey_2018} may be used to optimally adapt the pre-trained neural network to different high-level conditions. Also, clustering algorithms can be employed to distinguish driving styles and provide additional information to the neural network.

\section{Conclusions}
\label{section: CONCLUSIONS}
This paper presented a deep-learning framework to predict state trajectories of a human-driven vehicle, travelling on a track with complex geometry. It is based on a flexible problem formulation, so that the user can choose a generic set of signals to be forecast. To show a practical implementation of the algorithm, a predefined set of features useful for control applications was selected. One of the innovative contributions of this work is the inclusion of the road geometry into the prediction algorithm, to link the actions executed by the driver to the shape of the road that he/she travels. The prediction problem is solved with a deep neural network, whose complexity is constrained to allow for use in online control applications. The computational times obtained during tests confirmed the suitability of the method.

The efficacy of the proposed approach was validated in an interesting case study regarding motion cueing algorithms, with the corresponding dataset being generated in test sessions of a \mbox{non-professional} human driver on a dynamic driving simulator. The scheme showed good performance in predicting the desired signals, namely the longitudinal and lateral accelerations of the vehicle, and its yaw rate.

Finally, the generalisation capabilities of the trained neural network on two new test sets, generated by changing driver and track, respectively, were analysed and commented. Despite the new test sets were out of the scope of the training/validation sets, the neural network managed to forecast the main trends of the lateral acceleration and yaw rate. This demonstrated the effectiveness of the approach in linking the future predictions with the past dynamics of the driver-vehicle system and the road geometry. Future works shall define a minimal set of drivers and tracks to possibly improve the generalisation capabilities of the neural network on any new driver/track configuration. Also, the design of methods to enhance an automatic adaptation of the network to different drivers and tracks will be investigated. 

\bibliographystyle{IEEEtran}
\bibliography{IEEEabrv,root}

\begin{minipage}{0.30\columnwidth}
\includegraphics[width=0.99\columnwidth]{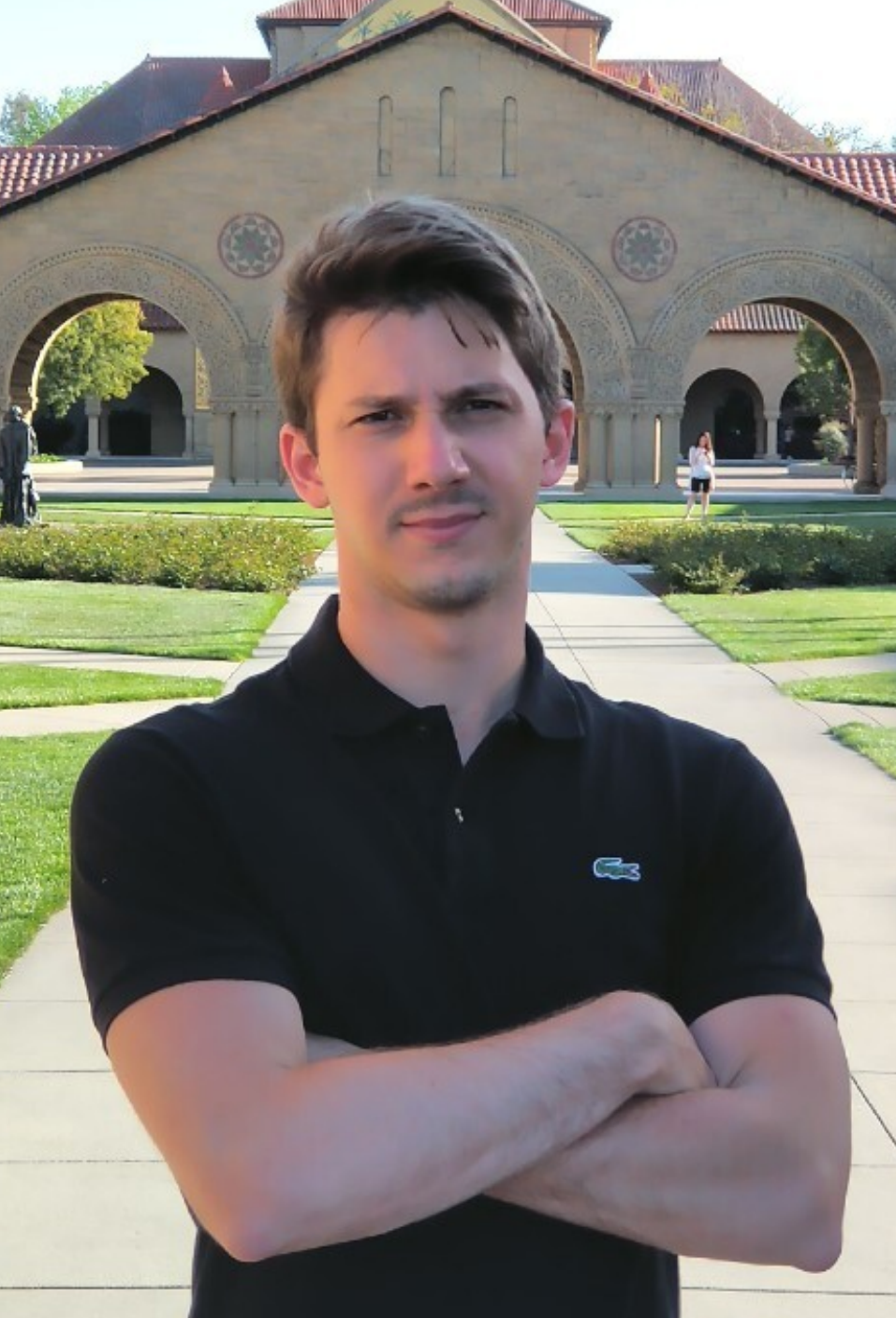}
\end{minipage} \hfill
\begin{minipage}{0.60\columnwidth}
\noindent \footnotesize{\textbf{Luca Paparusso} received the M.Sc. degree in mechanical engineering from Politecnico di Milano, Italy, in 2018, where he is pursuing the Ph.D. degree. He was research fellow at Istituto Italiano di Tecnologia (IIT), Italy, in 2019, and Visiting PhD at Stanford University, Autonomous Systems Laboratory (ASL), CA, USA, from 2021 to 2022. His research is focused on trajectory forecasting and motion control for efficient and safe autonomous navigation in multi-agent environments.}
\end{minipage}

\vspace*{2cm}
\begin{minipage}{0.30\columnwidth}
\includegraphics[width=0.99\columnwidth]{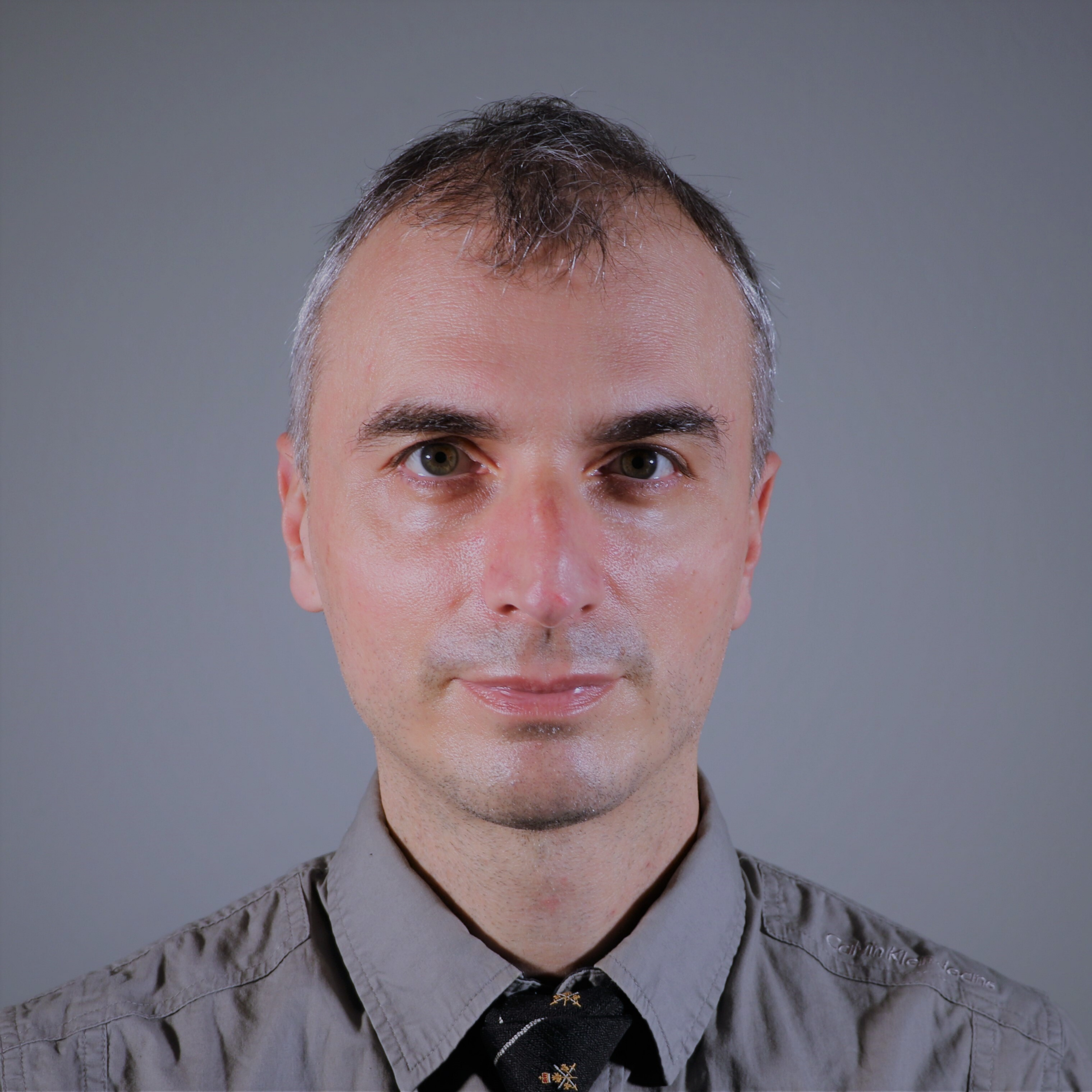}
\end{minipage} \hfill
\begin{minipage}{0.60\columnwidth}
\noindent \footnotesize{\textbf{Stefano Melzi} received the M.Sc. degree in mechanical engineering in 1999 and the Ph.D. in applied mechanics in 2003 from Politecnico di Milano, Italy, where he is associate Professor since 2014. His research activity deals in general with dynamics of mechanical systems and, in particular, on the dynamics of ground vehicles and interaction of wind with overhead transmission lines. He is author of nearly 150 publications almost entirely at international level.}
\end{minipage}

\vspace*{2cm}
\begin{minipage}{0.30\columnwidth}
\includegraphics[width=0.99\columnwidth]{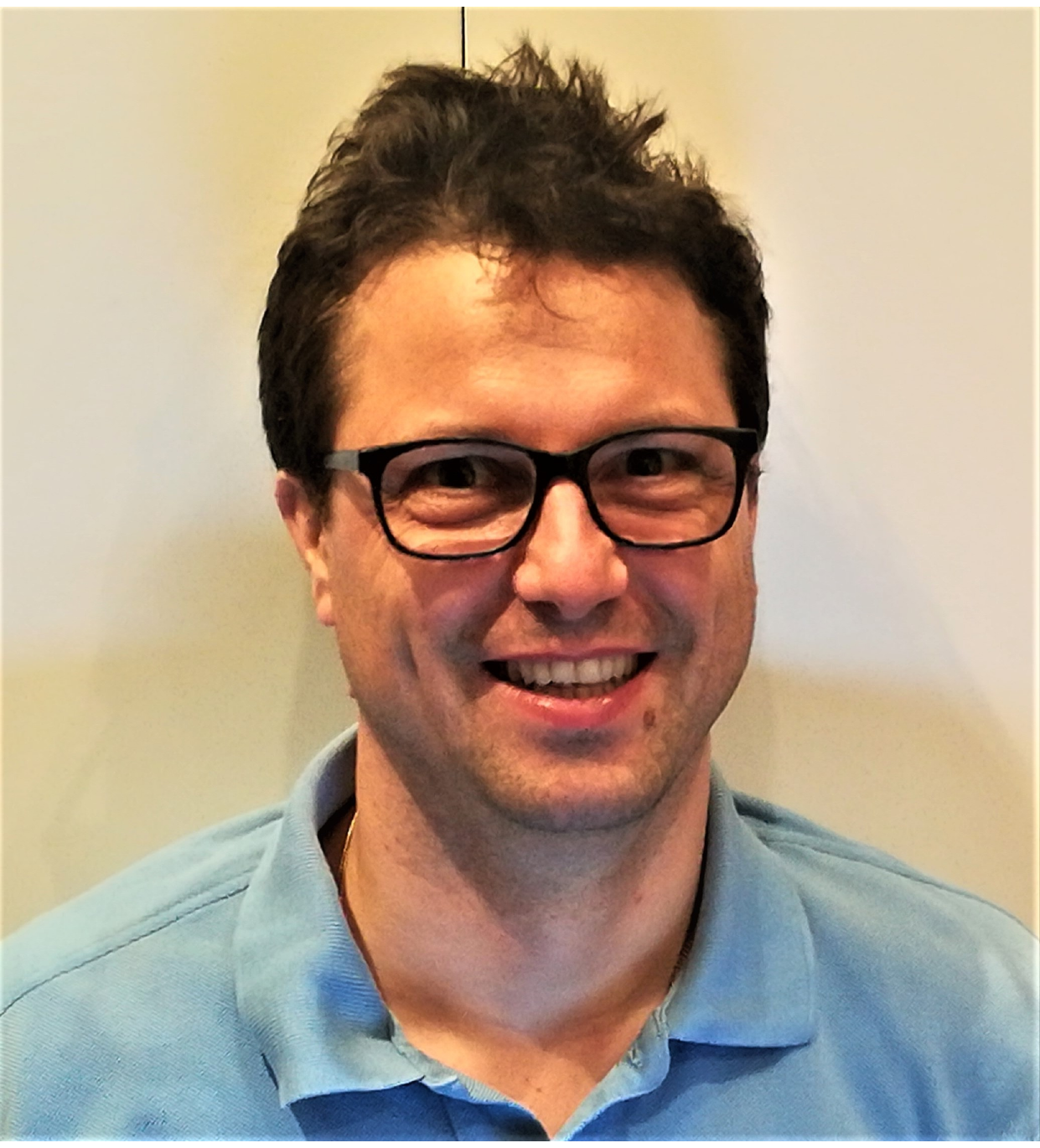}
\end{minipage} \hfill
\begin{minipage}{0.60\columnwidth}
\vspace*{0.9cm}
\noindent \footnotesize{\textbf{Francesco Braghin} received the M.Sc. degree in mechanical engineering in 1997 and Ph.D. in applied mechanics in 2001 from Politecnico di Milano, Italy, where he is full Professor since 2015. Author of more than 250 scientific publications, his research is carried out in the field of vehicle dynamics (road and railway) and mechatronics. In particular, as regards road vehicles, his research deals with the modelling of tires and their interaction with the soil, and the application of optimal control algorithms to the design of hybrid and electric vehicles, as well as to the development of fully autonomous vehicles.}
\end{minipage}

\end{document}